\documentclass[sigconf,nonacm]{acmart}
\AtBeginDocument{%
  }
\usepackage{multirow}
\usepackage{amsmath}
\usepackage{booktabs}
\usepackage{amsfonts}
\usepackage{makecell}
\setcopyright{none}
\acmISBN{978-1-4503-XXXX-X/2018/06}

\begin{document}

\title{Long Text to Predictive Features: LLM-Guided Blockwise Feature Engineering via Executable Program Search}

\author{Ziming Dai}
\authornote{Both authors contributed equally to this research.}
\affiliation{%
  \institution{City University of Hong Kong}
  \city{Hong Kong}
  \country{China}}
\email{phoenix.dai@my.cityu.edu.hk}

\author{Dabiao Ma}
\authornotemark[1]
\affiliation{%
  \institution{Qfin Holdings, Inc.}
  \city{Beijing}
  \country{China}
}
\email{madabiao-jk@qifu.com}

\author{Ziheng Guo}
\affiliation{%
  \institution{Tianjin University}
  \city{Tianjin}
  \country{China}}
\email{sygzh6@tju.edu.cn}

\author{Jack Dong}
\affiliation{%
  \institution{Carnegie Mellon University}
  \city{Pittsburgh}
  \country{USA}}
\email{jack.dong@bytedance.com}

\author{Zimu Zhou}
\correspondingauthor
\affiliation{%
  \institution{City University of Hong Kong}
  \city{Hong Kong}
  \country{China}}
\email{zimuzhou@cityu.edu.hk}

\renewcommand{\shortauthors}{Ziming Dai et al.}

\begin{abstract}
  Industrial risk-control systems typically rely on structured-data models for efficient prediction, yet substantial valuable information remains embedded in unstructured long text. Extracting this information through manual feature engineering is labor-intensive, while requiring a large language model (LLM) to process every real-time input may not meet practical deployment requirements. To address this challenge, we propose LLM-BlockFE, an LLM-guided offline feature construction framework that converts long text into executable feature programs, thereby avoiding LLM calls during online inference. LLM-BlockFE constructs feature programs by incrementally appending immutable code blocks and evaluates candidate features using a downstream model. To address the tendency of conventional greedy search to become trapped in suboptimal solutions, our method introduces a block-level rollback mechanism based on depth-calibrated credit allocation and advances multiple independent search trajectories in an interleaved manner, reducing redundant exploration by sharing fixed descriptions of each trajectory's exploration direction. After the search, the resulting programs are frozen and deployed to extract structured features for downstream prediction models. Across two public and two private datasets, LLM-BlockFE achieves absolute AUC improvements of 0.0069 to 0.0358 over the strongest baseline on each dataset in the full-dataset comparison. Post-launch monitoring across five deployed financial risk-control applications shows absolute KS improvements of 0.02 to 1.56 percentage points over the existing manually designed strategy.
\end{abstract}





\maketitle

\section{Introduction}
In high-stakes industrial applications such as financial risk control, predictive models must not only achieve high accuracy but also meet stringent constraints on inference latency and maintainability. Gradient-boosted tree models for structured data, such as XGBoost~\cite{chen2016xgboost} and LightGBM~\cite{ke2017lightgbm}, therefore remain widely used in industry because of their computational efficiency and seamless compatibility with existing business systems. Meanwhile, as business operations become more complex, unstructured long texts, such as transcripts of customer service calls, are increasingly becoming an important source of business information~\cite{jiang2023financial}. These texts contain rich information about user behavior, actual intent, and latent risk signals, but their unstructured form makes them difficult to use directly with commonly adopted structured-data models~\cite{fujii2022extraction}. Consequently, extracting effective features from raw long text has become a key bottleneck in industrial modeling.

At present, enterprises primarily rely on domain experts to design cleaning rules, text parsing logic, and feature computation programs, and then standardize the extracted outputs for use by downstream models. Variations such as colloquial language, omitted expressions, paraphrases, and cross-sentence semantics require manually designed parsing logic and feature rules to be continually extended and maintained. Existing text representation methods also have limitations: shallow features such as TF--IDF have limited capacity to represent contextual semantics. Dense vectors produced by pretrained models are difficult to interpret directly. Although long-sequence encoders can model long text context, their outputs typically require further processing to form structured feature columns with clear semantics~\cite{balek2025llm,beltagy2020longformer,zaheer2020big}. In addition, automated feature engineering methods that take existing structured feature columns as input primarily search for transformations and combinations of variables, without directly addressing feature construction from raw long text~\cite{zhang2023openfe}.

In recent years, large language models (LLMs) have demonstrated strong natural language understanding and code generation capabilities, offering a new technical approach to feature engineering~\cite{novikov2025alphaevolve,shojaee2025llm}. Some studies use LLMs to generate feature transformation code for existing tabular columns~\cite{abhyankar2025llm}, while others use LLMs to extract features with clear semantics from text~\cite{malberg2024felix}. For methods that rely on an LLM to generate feature values for individual samples, continuously arriving text inputs that require immediate scoring still need LLM inference, even when the feature definitions are established offline. This increases the latency and computational overhead of online feature extraction, placing pressure on industrial deployments with high concurrency.

To address these deployment constraints, we propose \textbf{LLM-BlockFE}, which formulates long text feature construction as the offline synthesis of executable feature programs. Unlike approaches that directly invoke an LLM for end-to-end prediction or per-sample feature extraction, LLM-BlockFE uses the LLM's semantic reasoning and code generation capabilities only during the offline stage to discover features and write the corresponding extraction programs. The resulting structured features are then provided to a downstream supervised model. This design allows the feature logic to be reviewed and maintained in code form and requires no LLM calls during the online stage.

Specifically, LLM-BlockFE constructs scalar feature extraction programs for long text and uses Incremental Program Search to generate a feature set whose members jointly improve downstream model performance. During the search, our method introduces a block-level rollback mechanism based on depth-calibrated credit allocation and advances multiple independent feature construction trajectories in an interleaved manner, reducing redundant exploration by sharing fixed descriptions of each trajectory's exploration direction. In deployment, the business system only needs to execute the finalized feature program code and its lightweight dependencies, with zero online LLM calls. Across five real-world financial risk-control applications, post-launch monitoring shows absolute KS improvements of 0.02 to 1.56 percentage points over the existing manually designed strategy across all 15 application-month comparisons. In addition, the offline evaluation results show that our method outperforms the strongest competing baseline on AUC values across all four datasets.

The contributions of this paper are summarized as follows:

\begin{itemize}
    \item We propose LLM-BlockFE for long text feature engineering, formulating feature construction as the offline synthesis of executable programs. We design a protocol for complete program generation and immutable block-level updates, providing explicit operational boundaries for incremental construction and local rollback. The resulting programs are used directly for online feature extraction, without requiring online LLM calls.

    \item We propose a depth-calibrated credit-guided backtracking mechanism that separately evaluates the current program's predictive performance and its opportunity for further expansion. Based on a node's recent expansion outcomes and historical baselines at the same depth, the mechanism determines whether to continue expansion or restore the parent program, reducing the dependence of single-path search on early choices.

    \item Experiments on multiple public and private datasets demonstrate that our method outperforms state-of-the-art (SOTA) baselines. Deployment and online evaluation in five real-world financial risk-control applications further validate the improvements that LLM-BlockFE brings to downstream predictive performance and its value for industrial applications.
\end{itemize}

\section{Related Work}

\textbf{Traditional Long Text Representation and Feature Engineering.} 
Traditional long text feature engineering typically transforms raw text into representations that can be processed by models. Common approaches include using TF-IDF to encode terms and their frequencies, extracting manually defined attributes as text statistics, and further compressing sparse term representations using dimensionality reduction methods such as SVD~\cite{qi2024comparison}. Another line of work uses pretrained language models to produce contextual representations. Long-sequence Transformers such as Longformer and BigBird extend the text lengths that models can process, and their encoded outputs can also be pooled to form document-level representations~\cite{beltagy2020longformer,zaheer2020big}.

Although the above methods can be used for text prediction, they remain insufficient to meet industrial requirements for feature controllability and interpretability. Traditional methods based on terms or manually defined rules have limited adaptability to variations in expression, such as colloquial language and omissions. Although pretrained language models can capture context, their outputs are typically high-dimensional dense vectors with no directly interpretable semantics, and require additional representation processing and inference computation. Therefore, the problem of automatically constructing compact, interpretable, and easily deployable structured features from raw long text remains unaddressed directly.

\noindent\textbf{Automated Feature Engineering.} 
Automated feature engineering typically starts from existing structured variables and searches for feature transformations that benefit downstream prediction. Methods such as OpenFE generate and select tabular features from a predefined operator space~\cite{zhang2023openfe}. Recent work further draws on the semantic knowledge of LLMs to propose candidate transformations. CAAFE generates feature engineering code based on dataset descriptions and table columns~\cite{hollmann2023large}, while OCTree converts predictive feedback expressed by a decision tree into natural-language reasoning to iteratively refine feature-generation rules~\cite{nam2024optimized}. LLM-FE formulates tabular feature transformation as program search and iteratively optimizes candidate programs through evolutionary-style search~\cite{abhyankar2025llm}. FAMOSE, in turn, adopts a ReAct-based agent workflow for feature discovery~\cite{burghardt2026famose}. Furthermore, NSR-Boost corrects the prediction residuals of industrial legacy models by gradually generating and verifying executable expert programs~\cite{dai2026nsr}.

However, these methods primarily search over existing tabular columns and their descriptions, with the central task of transforming or combining structured variables. This paper instead focuses on unstructured long text inputs: the system must construct features from the text content that can be used directly by tabular models, rather than merely transforming existing tabular columns.

\noindent \textbf{LLM-Based Feature Engineering for Unstructured Text.}
Recently, several studies have begun to explore feature engineering for long text using the semantic knowledge of LLMs. FELIX uses an LLM to generate interpretable features for text samples and applies them to downstream classification~\cite{malberg2024felix}. Balek \textit{et al.} study the extraction of a small number of semantic features from scientific literature, showing that these features can support text classification while remaining interpretable~\cite{balek2025llm}. Barlier \textit{et al.} investigate feature generation for joint prediction from text and tabular data~\cite{barlier2026llms}.

However, in these methods, candidate feature values still need to be extracted or assigned by an LLM for individual text samples. Batch processing or offline precomputation can mitigate this overhead, but for new samples that arrive continuously and require immediate scoring, feature extraction still incurs additional inference costs and latency, and creates a dependency on the serving infrastructure. In contrast, LLM-BlockFE uses an LLM to construct feature programs offline. After deployment, the system generates features by directly executing the frozen programs, without calling an LLM, thereby enabling fast inference.

\section{Notations and Problem Formulation}

Consider a supervised learning task with long text data, represented by the dataset:
\begin{equation}
\mathcal{D}=\{(u_i,\mathbf{s}_i,y_i)\}_{i=1}^{N},
\end{equation}
where $u_i\in\mathcal{U}$ denotes an unstructured long text input commonly encountered in industry, such as user dialogues, credit reports, application descriptions, or customer service records. $\mathbf{s}_i\in\mathbb{R}^{d_s}$ denotes optional pre-existing structured features, and $y_i\in\{0,1\}$ is a binary classification label. Although this paper primarily focuses on binary financial risk prediction tasks, the proposed feature engineering algorithm can also be extended to other supervised learning settings.

LLM-BlockFE aims to automatically construct compact and predictive structured features from long text, providing downstream models with effective information beyond that contained in pre-existing structured variables. Given text $u$ and structured features $\mathbf{s}$, the system transforms the text into a feature representation $\mathbf{z}(u)$ and makes predictions using the augmented input:
\begin{equation}
\hat{y}=g_{\theta}\left([\mathbf{s},\mathbf{z}(u)]\right).
\end{equation}
Here, $g_{\theta}$ is a supervised learning model that handles structured inputs. We use XGBoost as the primary downstream model for exposition and experiments.

Let $\mathcal{P}$ denote the space of executable text feature programs that satisfy predefined input--output interfaces and require no LLM calls at deployment. Any program $\Phi\in\mathcal{P}$ maps long text to a scalar feature, i.e., $\Phi:\mathcal{U}\rightarrow\mathbb{R}$. For a program set $\mathcal{F}=\{\Phi_1,\ldots,\Phi_m\}\subseteq\mathcal{P}$, the corresponding text feature vector is:
\begin{equation}
\mathbf{z}_{\mathcal{F}}(u)
=
[\Phi_1(u),\ldots,\Phi_m(u)].
\end{equation}
Below, $\theta(\mathcal{F})$ denotes the model parameters fitted using $[\mathbf{s},\mathbf{z}_{\mathcal{F}}(u)]$ under the prescribed training protocol, with $\theta(\Phi)$ used as shorthand for the single-program case. $\mathcal{M}$ denotes the predictive performance metric, for which we use ROC-AUC. $\mathcal{D}_{\mathrm{val}}$ is used for candidate comparisons during the search, while an independent test set is used for final performance evaluation. Detailed data usage is described in Section 4.

Program construction involves searching over both discrete code structures and numerical parameters. Moreover, the text signals extracted by different programs may be redundant or complementary~\cite{kim2024discovering}. We therefore describe the feature engineering objective at two levels: individual program construction and the combination of multiple programs.

\textbf{P1. Conditional text feature-program construction.}
Given pre-existing structured variables $\mathbf{s}$, the objective is to construct a new feature program $\Phi$ in the program space $\mathcal{P}$ that extracts predictive information from long text not yet covered by the current feature set:
\begin{equation}
\Phi^{*}
=
\arg\max_{\Phi\in\mathcal{P}}
\mathcal{M}\left(
g_{\theta(\Phi)}\left([\mathbf{s},\Phi(u)]\right);
\mathcal{D}_{\mathrm{val}}
\right).
\end{equation}
When the structured-feature baseline and evaluation protocol are fixed, maximizing this performance is equivalent to maximizing the predictive gain over the baseline. \textbf{P1} measures the predictive value of a single text feature program conditioned on pre-existing structured variables, without including text features from other trajectories in the evaluation.

\textbf{P2. Multi-program construction under a feature budget.}
Under a finite feature budget, the objective is to construct multiple text feature programs whose outputs achieve strong downstream predictive performance when used jointly with pre-existing structured variables:
\begin{equation}
\mathcal{F}^{*}
=
\arg\max_{
\substack{
\mathcal{F}\subseteq\mathcal{P}\\
|\mathcal{F}|\leq K_{\max}
}
}
\mathcal{M}\left(
g_{\theta(\mathcal F)}\left(
[\mathbf{s},\mathbf{z}_{\mathcal{F}}(u)]
\right);
\mathcal{D}_{\mathrm{val}}
\right).
\end{equation}
Here, $K_{\max}$ is the maximum number of text feature programs retained and deployed, constraining the dimensionality of the added features. Since a program's marginal value depends on the information extracted by other programs in the set, combining programs obtained by independently solving \textbf{P1} does not necessarily yield the optimal set for \textbf{P2}.

LLM-BlockFE uses \textbf{P1} as the independent search objective for each trajectory. Fixed exploration directions and shared direction descriptions encourage the exploration of different text signals. After the search, the system combines the best programs archived by the individual trajectories and trains the final downstream model. This procedure approximates \textbf{P2}: direction sharing reduces redundant exploration, while the actual complementarity among programs and the gains from their combination are evaluated empirically.

\section{Methodology}
\begin{figure*}[t]
    \includegraphics[width=\textwidth]{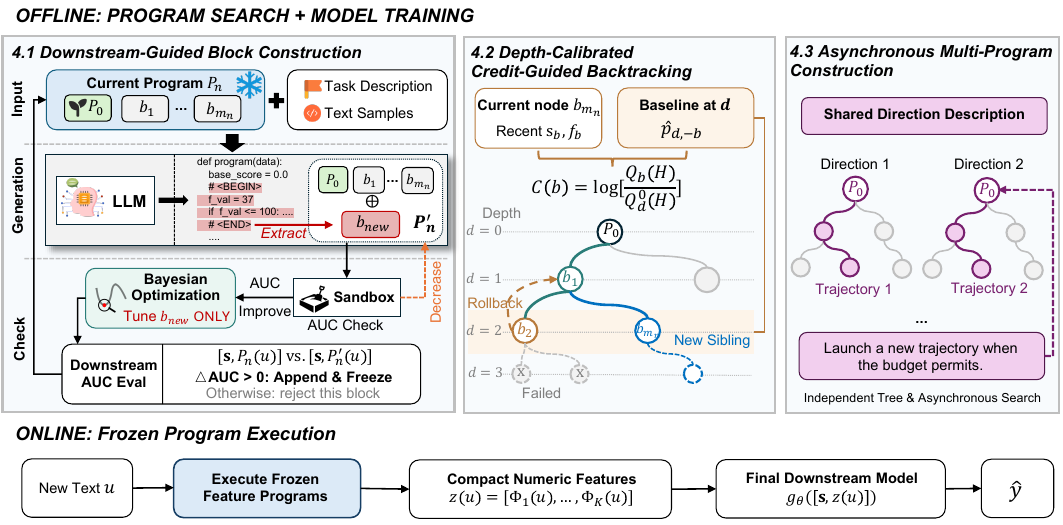}  
    \caption{Overview of LLM-BlockFE, which searches for text feature programs offline and executes the selected programs for online prediction.}
    \label{fig:framework}
\end{figure*}

As shown in Figure~\ref{fig:framework}, the LLM-BlockFE architecture consists of three tightly coupled components. First, the system generates a complete candidate program and extracts the added code block using predefined markers, enabling incremental updates while keeping existing code unchanged. Parameters within the added block are then locally tuned through Bayesian optimization. Second, a depth-calibrated credit mechanism evaluates the relative opportunity of obtaining an acceptable child through further expansion of the current program node, guiding continued expansion or dynamic backtracking. Finally, guided by fixed exploration directions, the system advances multiple search trajectories asynchronously in an interleaved manner and shares direction descriptions to reduce redundant exploration. After the search, the system collects and freezes the best programs from all trajectories for online feature extraction, without requiring online LLM calls.

\subsection{Downstream-Guided Block Construction}

To construct a single text feature program as defined in Section 3, LLM-BlockFE starts from the Seed Program $P_0$, which specifies the input, output, and code insertion point. $P_0$ defines only the available libraries and the standardized output. Consider the $n$-th search trajectory. The current program consists of the initial program and the code blocks accepted along the current path:
\begin{equation}
P_n=P_0\oplus b_{n,1}\oplus\cdots\oplus b_{n,m_n},
\end{equation}
where $\oplus$ is appending a code block at the designated location, and $m_n$ is the number of code blocks on the current path. The program maps a long text input $u$ to a scalar feature, i.e., $\Phi_n(u)=P_n(u)$.

\begin{figure}[t]
    \includegraphics[width=\linewidth]{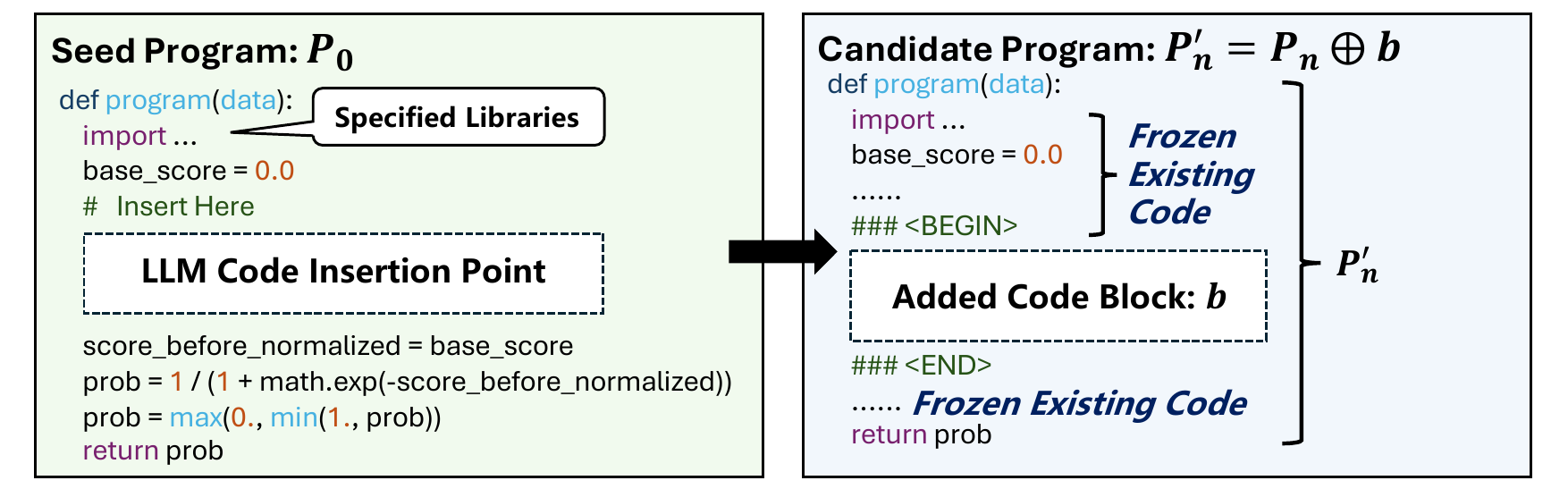}  
    \caption{Illustration of the Seed program and LLM-generated candidate program.}
    \label{fig:python}
\end{figure}

To maintain coherence in the overall program logic while allowing local updates to be reversed, LLM-BlockFE adopts a program update mechanism based on ``complete generation and incremental extraction'', as shown in Figure~\ref{fig:python}. Directly requesting an isolated code addition from the LLM can readily lead to mismatches, such as variable name conflicts, calls to undefined interfaces, or semantic discontinuities with existing code, because of the model's limited awareness of the surrounding program. LLM-BlockFE therefore uses the complete program as the unit of generation. The LLM receives $P_n$, a very small set of text samples $\mathcal{D}_{\mathrm{llm}}\subset\mathcal{D}_{\mathrm{train}}$, the task description, callable tools, and previous generation summaries. It generates complete candidate code, marks the added region using predefined markers (\texttt{<BEGIN>...<END>}), and provides a summary of the current round for subsequent iterations. The system then precisely extracts the added code block $b$ using a regular expression, constructs the candidate program $P_n'=P_n\oplus b$, and strictly checks whether the existing code outside the markers has changed. Any candidate that impermissibly modifies the original code is rejected. Generating with the full program context promotes syntactic and logical compatibility between the new and existing code, while block-level extraction preserves clear boundaries between them and provides the basis for subsequent local backtracking.

The candidate program first undergoes an execution check in a restricted environment. Candidates that fail to execute or produce invalid outputs are immediately discarded. For valid candidates, the system compares the AUCs of the text features produced by $P_n'$ and $P_n$. If the untuned candidate decreases this AUC, it is discarded early. This step serves as a low-cost heuristic screening procedure to reduce the overhead of subsequent parameter optimization.

For candidates that pass the initial screening and contain tunable parameters, the LLM identifies the parameters to optimize, their search ranges, and the optimization objective based on the added block. Because operations such as threshold tests make usable gradients unavailable for the program output and its AUC with respect to the parameters, the system uses Bayesian optimization and writes the resulting parameter values back into the added block. Optimization affects only the added block. The existing code and its frozen parameters remain unchanged. The program with the updated parameter values must pass the execution check again and is still denoted by $P_n'$ below.

Finally, the system uses a lightweight XGBoost model with a fixed configuration to evaluate the candidate's downstream value. To prevent data leakage and overfitting during the search, the evaluation procedure strictly separates data usage: the lightweight XGBoost model is fitted only on a training subset $\mathcal{D}_{\mathrm{fit}}\subset\mathcal{D}_{\mathrm{train}}$, while performance is evaluated on an independent validation set $\mathcal{D}_{\mathrm{val}}$. The system uses $[\mathbf{s},P_n(u)]$ and $[\mathbf{s},P_n'(u)]$, respectively, as inputs to fit the evaluation models $g_{\theta(P_n)}$ and $g_{\theta(P_n')}$ on $\mathcal{D}_{\mathrm{fit}}$, and then computes their AUCs on $\mathcal{D}_{\mathrm{val}}$. The conditional gain of the added block is defined as:
\begin{equation}
\begin{aligned}
\Delta_{n,b}
={}&
\mathcal{M}\left(
g_{\theta(P_n')}\left([\mathbf{s},P_n'(u)]\right);
\mathcal{D}_{\mathrm{val}}
\right)\\
&-
\mathcal{M}\left(
g_{\theta(P_n)}\left([\mathbf{s},P_n(u)]\right);
\mathcal{D}_{\mathrm{val}}
\right),
\end{aligned}
\end{equation}
 where $\mathcal{M}$ denotes AUC. For a new trajectory in which no code blocks have yet been accepted, the parent model uses only the structured features $\mathbf{s}$. When $\Delta_{n,b} > 0$, the system accepts and freezes the added block and appends it to the current path.

Once feature extraction by all programs is complete, the resulting feature matrix is combined with the structured features to train the final downstream risk-control model on $\mathcal{D}_{\mathrm{down}}\subseteq\mathcal{D}_{\mathrm{train}}$. Its generalization performance is then evaluated on an independent test set $\mathcal{D}_{\mathrm{test}}$.

\subsection{Depth-Calibrated Credit-Guided Backtracking}

The previous subsection introduced candidate code-block construction, but the append-only constraint does not, by itself, determine how the program space is explored. If the search always expands along the current path, blocks accepted early may constrain subsequent exploration. NSR-Boost uses an annealed threshold to allow small performance decreases early in the search, but still uses the current program as the starting point for subsequent iterations~\cite{dai2026nsr}. LLM-BlockFE introduces depth-calibrated credit-guided backtracking, using the current node's expansion opportunity to determine whether to continue the search or restore the parent program to explore an alternative branch~\cite{yao2023tree,parsert2024reinforcement}.

Specifically, each search trajectory can be represented as a search tree rooted at $P_0$. Each non-root node is uniquely identified by its terminal code block $b$ and corresponds to the complete program formed by accumulating code blocks from the root. The node depth $d$ is defined as the number of accepted code blocks on its path. A node's downstream AUC reflects the predictive performance of the current program, while its credit measures the likelihood of obtaining a valid child through further expansion from that node. For the $t$-th eligible expansion attempt at node $b$, we assign $y_{b,t}=1$ if the candidate passes all predictive gates and is accepted, and $y_{b,t}=0$ if it fails the initial text AUC screening, local parameter optimization, or the downstream AUC gate. Non-predictive failures, such as syntax or execution errors, are excluded from these statistics.

Because the difficulty of adding code varies substantially across depths, we construct a baseline from historical observations at the same depth~\cite{fialho2010toward}. Let block $b$ have depth $d$, and let $S_{-b}$ and $F_{-b}$ denote the lifetime numbers of successes and failures, respectively, among all other non-root blocks. The global success-rate estimate is:
\begin{equation}
\bar{p}_{-b}
=
\begin{cases}
p_0, \quad S_{-b}+F_{-b}=0,\\
\dfrac{S_{-b}+\frac{1}{2}}{S_{-b}+F_{-b}+1},  \quad \text{otherwise},
\end{cases}
\end{equation}
where $p_0$ is a fixed cold-start success rate, and Jeffreys smoothing is used when historical observations are available~\cite{garthwaite2024locally}. Further, let $S_{d,-b}$ and $F_{d,-b}$ denote the lifetime numbers of successes and failures, respectively, among other blocks at the same depth. The depth-specific success rate is estimated using empirical-Bayes partial pooling as~\cite{agarwal2022hierarchical}:
\begin{equation}
\hat{p}_{d,-b}
=
\frac{S_{d,-b}+\kappa\bar{p}_{-b}}
{S_{d,-b}+F_{d,-b}+\kappa},
\end{equation}
where $\kappa$ controls the degree of shrinkage of within-depth observations toward the global baseline. The performance of block $b$ is modeled using recent valid attempts to adapt to changes in candidate quality and the remaining search space during the search. Let $s_b$ and $f_b$ denote the numbers of successes and failures, respectively, for block $b$ in the recent window. Under a Beta-Bernoulli model, its Beta posterior parameters are defined as:
\begin{equation}
\alpha_b=\kappa\hat{p}_{d,-b}+s_b,
\qquad
\beta_b=\kappa(1-\hat{p}_{d,-b})+f_b.
\end{equation}

Observations outside the recent window are retained in the lifetime statistics and used to estimate the global and depth-specific baselines for other blocks. Let $H$ denote the short-term prediction horizon. Based on the posterior distribution of block $b$, its predicted probability of producing at least one valid child in the next $H$ direct attempts is:
\begin{equation}
Q_b(H)
=
1-\prod_{i=0}^{H-1}
\frac{\beta_b+i}{\alpha_b+\beta_b+i}.
\end{equation}
For a block at the same depth with no observations of its own, this probability is:
\begin{equation}
Q_d^0(H)
=
1-\prod_{i=0}^{H-1}
\frac{\kappa(1-\hat{p}_{d,-b})+i}{\kappa+i}.
\end{equation}

LLM-BlockFE defines block credit as the log ratio of a block's expansion opportunity to the baseline at the same depth:
\begin{equation}
C(b)=\log R_b,
\qquad
R_b=\frac{Q_b(H)}{Q_d^0(H)}.
\end{equation}
Thus, $C(b)=0$ indicates that the block's opportunity matches the same-depth baseline. $C(b)>0$ indicates an opportunity above the baseline. $C(b)<0$ indicates an opportunity below it. A new block has no observations of its own and is therefore initialized with $C(b)=0$. A successful block updates only the posterior of its direct parent.

When the relative opportunity $R_b$ of the current block $b$ falls below the threshold $\rho$, LLM-BlockFE rolls back one terminal block. After rollback, the search resumes from the restored parent and explores a new sibling. Since the root has no block to roll back, it is not assigned ordinary credit. 

\subsection{Asynchronous Multi-Program Construction}

To construct multiple text feature programs, LLM-BlockFE advances multiple search trajectories in an interleaved manner. Each trajectory starts from the initial program $P_0$ and separately maintains its current search path, code-block history, rollback state, and best-program archive. Once a trajectory's first code block is accepted, the LLM establishes a fixed exploration direction based on the feature semantics of that block. The system can then initiate new trajectories when the budget permits, without waiting for existing trajectories to complete their searches.

Trajectories share only descriptions of their exploration directions. When generating a candidate code block, the LLM continues feature construction based on the current program and fixed direction of its own trajectory, while referring to the direction descriptions of other trajectories to avoid repeatedly exploring the same targets~\cite{han2025tabular}. The program code, best paths, and feature outputs of other trajectories are excluded from the generation context of the current trajectory. The fixed direction constrains the primary exploration target but allows different trajectories to use the same keywords or tools.

Each trajectory follows the candidate construction and evaluation procedure in Section 4.1, independently performing initial text AUC screening, local Bayesian optimization, and downstream AUC evaluation. Downstream evaluation uses only the pre-existing structured features and the program output of the current trajectory, excluding features from other trajectories. Accepted code blocks are added to the current path. When the resulting downstream AUC exceeds the trajectory's historical maximum, its best-program archive is updated. Rollback changes only the current path. It neither deletes the archived best program nor changes the fixed exploration direction.

Candidate generation and evaluation can run concurrently across trajectories, while candidate processing and state updates within each trajectory proceed sequentially. After the search, the system freezes the best programs saved by all trajectories and jointly trains the final downstream model using their outputs together with the pre-existing structured variables.

\section{Experiments}

\begin{table*}[t]
    \centering
    \small
    \setlength{\tabcolsep}{4pt}
    \begin{tabular}{l c c c c c c c c c c c}
        \toprule
        \makecell[c]{\textbf{Dataset}} &
        \makecell[c]{\textbf{Base-XGB}} &
        \makecell[c]{\textbf{TextStats}\\\textbf{+XGB}} &
        \makecell[c]{\textbf{TF-IDF}\\\textbf{+LR}} &
        \makecell[c]{\textbf{TF-IDF}\\\textbf{+XGB}} &
        \makecell[c]{\textbf{TF-IDF+SVD}\\\textbf{+XGB}} &
        \makecell[c]{\textbf{Embedding}\\\textbf{+XGB}} &
        \makecell[c]{\textbf{OpenFE}} &
        \makecell[c]{\textbf{CAAFE}} &
        \makecell[c]{\textbf{OCTree}} &
        \makecell[c]{\textbf{LLM-FE}} &
        \makecell[c]{\textbf{Ours}} \\
        \midrule
        Remark
        & 0.6265 & 0.6146 & 0.6058 & 0.6323 & 0.6261 & 0.5685
        & 0.6439 & 0.6314 & 0.6435 & \underline{0.6446} & \textbf{0.6804} \\
        Report
        & 0.5323 & 0.5245 & 0.5232 & 0.5314 & 0.5227 & 0.5125
        & 0.5250 & \underline{0.5336} & 0.5159 & 0.5324 & \textbf{0.5545} \\
        CFPB-Timely
        & 0.8571 & 0.8418 & 0.8699 & 0.8663 & 0.8707 & 0.8717
        & 0.8541 & 0.8684 & 0.8701 & \underline{0.8729} & \textbf{0.8881} \\
        GossipCop
        & 0.9145 & 0.9224 & 0.8603 & 0.9380 & \underline{0.9403} & 0.9330
        & 0.9206 & 0.9257 & 0.9206 & 0.9242 & \textbf{0.9472} \\
        \bottomrule
    \end{tabular}
    \caption{AUC comparison across datasets. Best results are highlighted in bold, and second-best results are underlined.}
    \label{tab:auc_comparison}
\end{table*}

\subsection{Experimental Setup}

\textbf{Datasets.} We evaluate the proposed method on two public and two private datasets, using models trained only on structured attributes as baselines to measure the incremental gains from text features. The public datasets include CFPB-Timely, constructed from the CFPB Consumer Complaint Database~\cite{cfpb_consumer_complaints}, with complaint narratives as text inputs and whether the company responded in a timely manner as the prediction target. It contains 200,000 samples and seven structured features. The other public dataset is a news veracity dataset constructed from the GossipCop subset of FakeNewsNet~\cite{shu2020fakenewsnet}, with news article text as input alongside structured features such as the number of images. The private datasets, Remark and Report, originate from Qfin Holdings' credit risk-control applications and contain de-identified text records with corresponding risk labels. Remark comprises remarks made by contacts about target users, whereas Report comprises call records between customer-service agents and users.

\noindent \textbf{Evaluation protocol.} We use AUC to evaluate predictive performance. Each dataset is divided into 60\% training, 20\% validation, and 20\% test sets. We conduct independent experiments using five random seeds and report the mean results.

\noindent \textbf{Baselines.} Conventional text representation methods include TF--IDF, TF--IDF + SVD, statistical text features (TextStats), and dense embeddings (Embedding). Automated feature engineering methods include OpenFE~\cite{zhang2023openfe}, CAAFE~\cite{hollmann2023large}, OCTree~\cite{nam2024optimized}, and LLM-FE~\cite{abhyankar2025llm}. During feature engineering, OCTree uses TextStats extracted from long text as input, while LLM-FE and CAAFE take long text directly as input. We also include FELIX~\cite{malberg2024felix} as an LLM-based text feature extraction baseline. Due to the cost of per-sample LLM calls, we compare its performance and latency only on subsets of Remark and CFPB-Timely. Except for an additional TF--IDF + Logistic Regression (LR) comparison, all baselines use XGBoost with hyperparameters optimized via Bayesian optimization as the downstream predictor.

\noindent \textbf{Implementation settings.} Each baseline follows its original procedure. LLM-BlockFE and all LLM-based baselines use the locally deployed Seed-OSS-36B-Instruct model~\cite{seed2025seed-oss} with a sampling temperature of $t=0.1$. LLM-BlockFE uses $K=5$ search trajectories, provides 20 samples to the LLM per round, and sets the maximum search depth to 5. After the search, the best programs saved by the individual trajectories are frozen for feature extraction. 
Further dataset details and hyperparameter settings are provided in Appendices~\ref{apx:A.1} and~\ref{apx:a2}.

\subsection{Offline Performance and Efficiency}

Table~\ref{tab:auc_comparison} summarizes the classification performance of LLM-BlockFE and the baseline methods across four datasets. Except for TF--IDF + LR, which uses logistic regression, all feature engineering baselines use XGBoost as the downstream prediction model. LLM-BlockFE achieves the best performance on all datasets, with AUCs of 0.6804, 0.5545, 0.8881, and 0.9472 on the private datasets Remark and Report and the public datasets CFPB-Timely and GossipCop, respectively. These correspond to improvements of 0.0358, 0.0209, 0.0152, and 0.0069 over the strongest baseline on each dataset. These results confirm the method's generalization ability and superior performance on both private data and general public long text tasks.

Comparing the different baseline paradigms reveals that LLM-based feature engineering methods significantly outperform conventional statistical methods and traditional AutoFE methods overall on the private risk-control datasets, where noise is greater and informative signals are more subtle and complex. This finding clearly demonstrates the stronger capabilities of LLMs to extract deeper semantics and uncover implicit associations in noisy long text. Among the LLM-based methods, LLM-BlockFE further achieves the best performance across all datasets through incremental code-block evolution and credit-guided dynamic backtracking. Moreover, unlike methods that rely on real-time online LLM calls, LLM-BlockFE moves the entire process of constructing complex feature programs to the offline stage. Online deployment requires only sequential execution of the frozen feature code, incurring zero computational or latency overhead from online LLM calls.

\begin{figure}[t]
    \includegraphics[width=\linewidth]{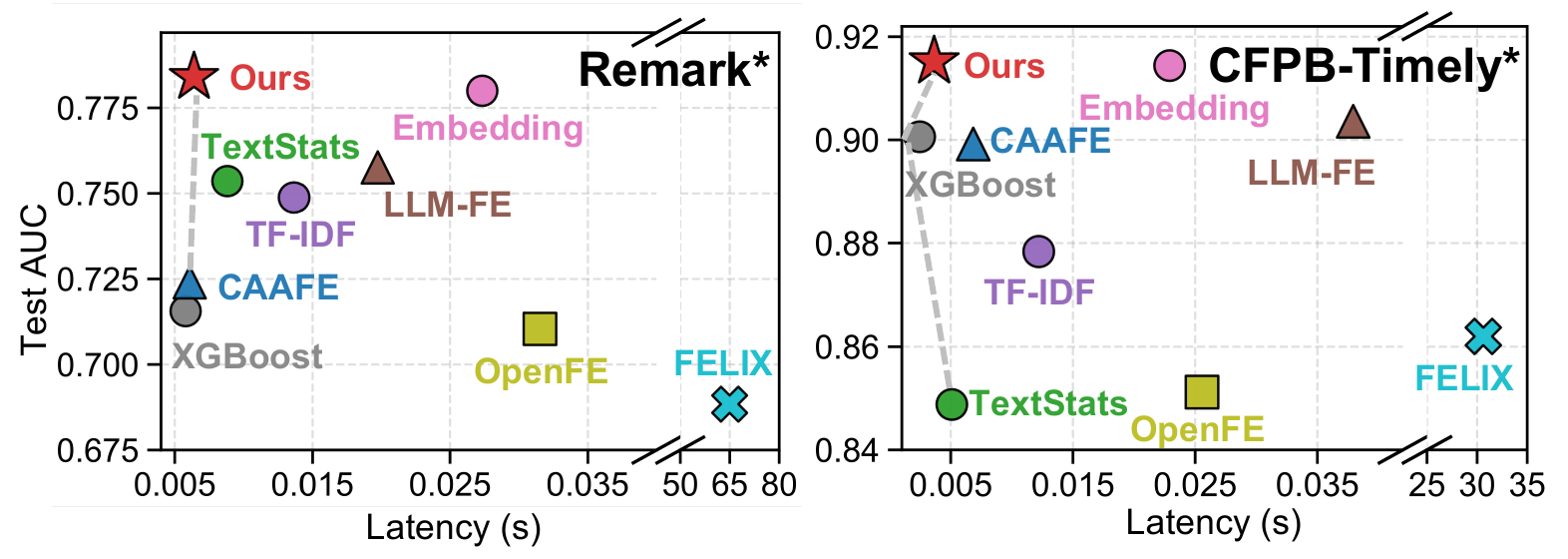}  
    \caption{Inference latency and predictive performance across methods.}
    \label{fig:efficiency}
\end{figure}

Figure~\ref{fig:efficiency} compares the classification performance and end-to-end latency of the methods during online inference. To precisely assess inference efficiency, we randomly sample 100 instances from each full test set to construct the latency benchmark subsets Remark$^{*}$ and CFPB-Timely$^{*}$. LLM-BlockFE achieves the highest AUC on both benchmark subsets
while maintaining millisecond-level inference latency.

Compared with FELIX, a strong baseline that also focuses on efficiency, LLM-BlockFE achieves higher AUC while reducing inference latency by orders of magnitude. This substantial advantage stems from complete architectural decoupling: LLM-BlockFE confines the LLM's complex logical understanding and feature search processes entirely to the offline stage. Online deployment requires only sequential execution of the frozen programs for feature extraction, eliminating the substantial computational overhead and latency bottleneck of online LLM inference for individual samples.

\subsection{Production Deployment and Post-Launch Impact}

\begin{figure*}[t]
    \includegraphics[width=\textwidth]{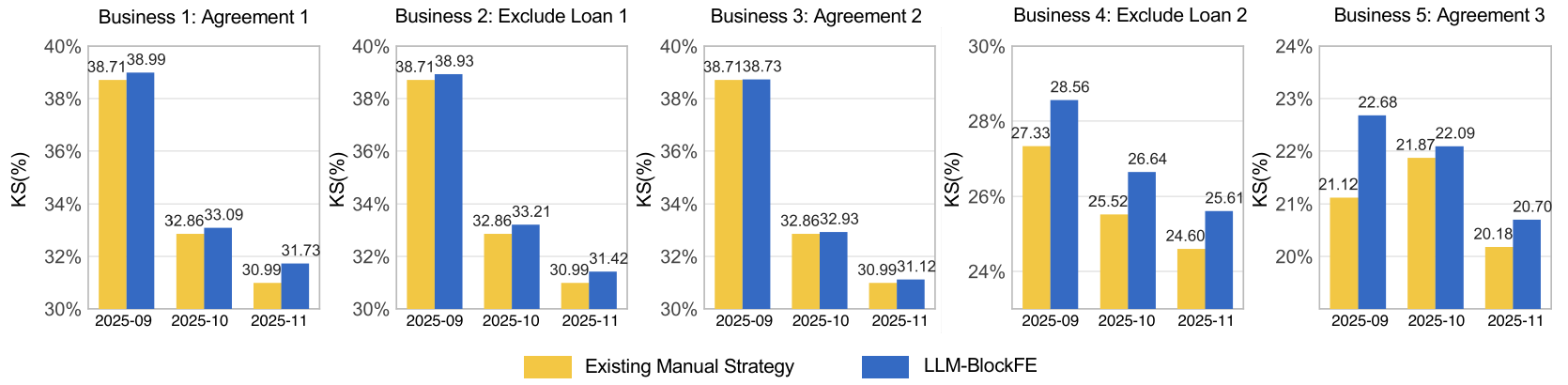}  
    \caption{Online performance of LLM-BlockFE across five production risk-control applications. The figure compares monthly KS values for LLM-BlockFE and the existing manually designed strategy on the same online samples from September to November 2025, covering 2,694,342 real-world users.}
    \label{fig:online}
\end{figure*}

LLM-BlockFE has been deployed in five real-world financial risk-control applications. During online inference, the system executes feature programs generated and frozen offline, without calling an LLM. Because this feature engineering process does not alter the original structured data, the production system can compute results for two approaches in parallel on exactly the same online samples: one retains the existing manually designed strategy, while the other uses text features generated by LLM-BlockFE. Both approaches use the same samples, labels, structured features, downstream model, and other business strategies. Their performance differences therefore reflect the impact of changing the text feature approach.

Figure~\ref{fig:online} shows the monthly KS values of both approaches across the five applications from September to November 2025, covering 2,694,342 real users. Across all 15 application--month comparisons, the LLM-BlockFE approach achieves higher KS than the existing manually designed strategy, with absolute improvements of 0.02--1.56 percentage points. These production results show that, with all other inputs and modeling procedures held fixed, incorporating text features generated by LLM-BlockFE yields higher discrimination in all five risk-control applications.

The system's practical business value is further validated through real-world online A/B testing. In an online evaluation covering approximately RMB 1 billion in loan disbursements, the treatment group significantly reduced the bad debt rate from 8.0\% to 7.7\% while keeping the number of approved applicants unchanged. The company has developed more than \textbf{200} effective feature chains using this framework, achieving comprehensive coverage of nearly all customer segments. From an engineering perspective, feedback indicates that the deployed feature computation pipelines incur very low online overhead and require no GPUs, demonstrating strong potential for broader industrial adoption.

\subsection{Ablation Study}
\begin{table}[t]
    \centering
    \small
    \setlength{\tabcolsep}{7pt}
    \begin{tabular}{l c c}
        \toprule
        \textbf{Setting} &
        \textbf{AUC} &
        \textbf{AUC Drop} \\
        \midrule
        Base-XGB                       & 0.6190 & 0.0570 \\
        w/o Append-only Constraint         & 0.6357 & 0.0403 \\
        w/o Bayesian Optimization          & 0.6383 & 0.0377 \\
        w/o Downstream AUC Gate            & 0.6268 & 0.0492 \\
        w/o Backtracking                   & 0.6470 & 0.0290 \\
        w/o Credit Guidance                & 0.6573 & 0.0187 \\
        w/o Depth Calibration              & 0.6640 & 0.0120 \\
        w/o Cross-Program Sharing          & 0.6708 & 0.0052 \\
        \textbf{Full Method}               & \textbf{0.6760} & 0.0000 \\
        \bottomrule
    \end{tabular}
    \caption{Ablation study on the Remark dataset. AUC Drop denotes the absolute performance decrease relative to the full method.}
    \label{tab:ablation_study}
\end{table}

Table~\ref{tab:ablation_study} summarizes the ablation results on Remark. Each setting was evaluated over a separate set of five independent runs, distinct from those used for the main comparison in Table~\ref{tab:auc_comparison}. The reported AUC is the mean across these runs. The complete method achieves an AUC of 0.6760, an absolute improvement of 0.0570 over Base-XGB. Removing any module reduces AUC, indicating that each design contributes to the final performance under the current experimental setting. Removing the downstream AUC gate has the largest impact: AUC drops to 0.6268, which is 0.0492 below the complete method and only 0.0078 above Base-XGB. This shows that even if a candidate program executes successfully and passes the initial text AUC screening and local parameter optimization, it may not provide a conditional gain to the downstream model. Evaluating the downstream AUC gain for each block helps filter out candidate programs that fail to improve predictive performance.

Ablations of the program construction and search mechanisms also demonstrate the roles of their respective components. Removing the append-only constraint reduces AUC to 0.6357, suggesting that restricting the scope of changes in each round and freezing accepted code blocks helps stabilize program construction. Removing Bayesian optimization reduces AUC to 0.6383, demonstrating the contribution of local parameter optimization to final predictive performance. For the search mechanism, disabling backtracking reduces AUC to 0.6470, indicating that unidirectional expansion is susceptible to constraints imposed by early choices. When credit guidance is removed, the system instead backtracks after a fixed threshold of three failures, and AUC drops to 0.6573. Removing depth calibration reduces AUC to 0.6640. These results indicate that credit guidance provides more effective backtracking decisions than a fixed failure threshold, while depth calibration further improves search outcomes on this dataset. Removing cross-program sharing yields an AUC of 0.6708. Although the decrease is relatively small, it still indicates an additional benefit from sharing exploration directions.

\subsection{Generalization Analysis}

\begin{figure}[t]
    \includegraphics[width=\linewidth]{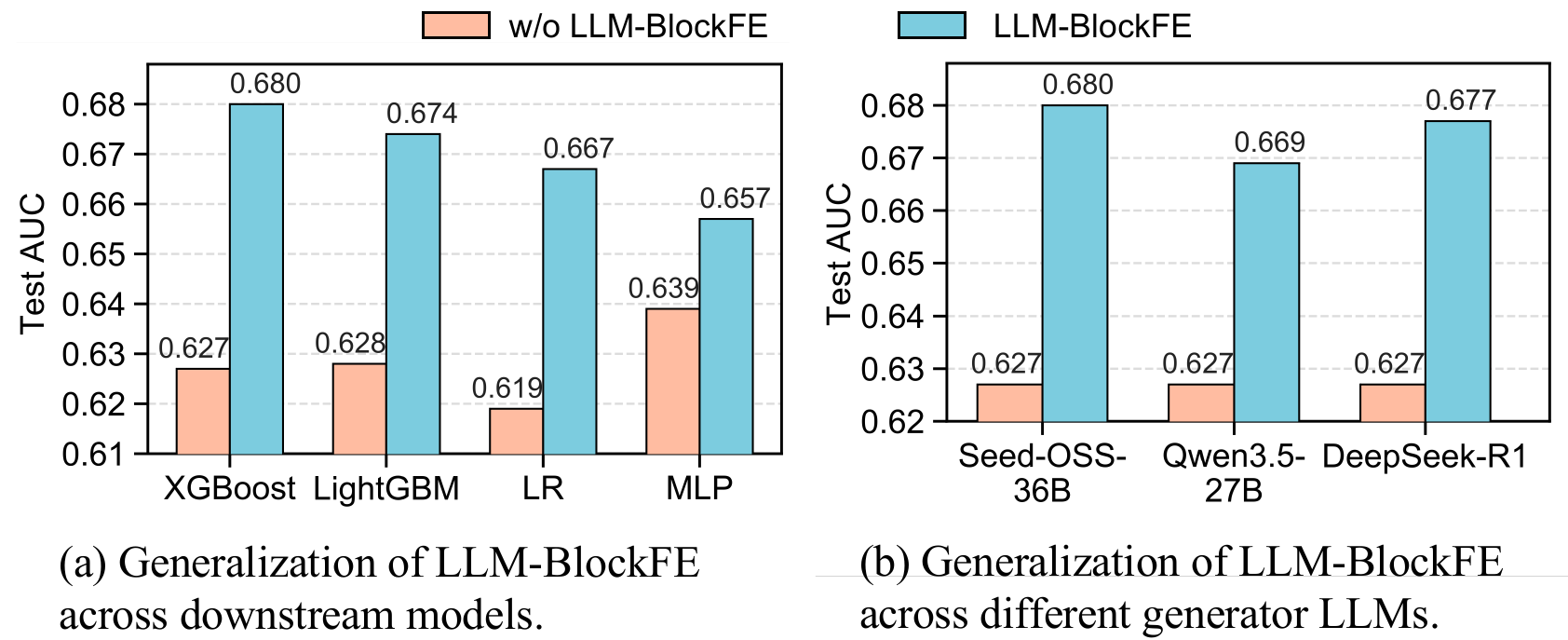}  
    \caption{Generalization of LLM-BlockFE across downstream models and generator LLMs on the Remark dataset.}
    \label{fig:downstream}
\end{figure}

\textbf{Generalization across downstream models.} Figure~\ref{fig:downstream}(a) examines whether the features generated by LLM-BlockFE are effective across different downstream models. Incorporating the generated features improves test AUC for all four models. XGBoost and LightGBM achieve gains of 0.053 and 0.046, respectively, while logistic regression and MLP improve by 0.048 and 0.018. These gains span tree-based, linear, and neural models, suggesting that the generated features provide useful predictive information beyond a single model architecture. The variation in gains also indicates that downstream models differ in how effectively they exploit the generated features.

\noindent\textbf{Generalization across feature-generating LLMs.} Figure~\ref{fig:downstream}(b) evaluates the sensitivity of LLM-BlockFE to the choice of feature-generating LLM. Relative to the baseline AUC of 0.627 without LLM-BlockFE, using Seed-OSS-36B-Instruct, Qwen3.5-27B~\cite{qwen3.5}, and DeepSeek-R1~\cite{deepseekai2025deepseekr1incentivizingreasoningcapability} yields test AUCs of 0.680, 0.669, and 0.677, respectively, corresponding to improvements of 0.053, 0.042, and 0.050. All three generation models improve performance, indicating that the method’s effectiveness does not depend on a single generation model. The differences in gain magnitude nevertheless show that generator choice affects the final predictive performance.

\subsection{Feature Contribution and Case Studies}

To analyze the role of the textual features generated by LLM-BlockFE in downstream model predictions, Table~\ref{tab:feature_contribution} summarizes their mean absolute SHAP values and relative contribution shares~\cite{lundberg2017unified}. On the private datasets Remark and Report, the generated textual features account for 41.4\% and 61.1\% of the total attribution, respectively. On the public datasets CFPB-Timely and GossipCop, the corresponding shares are 9.5\% and 19.4\%. These results indicate that, under the current experimental settings, the generated textual features account for substantial attribution shares in model predictions on the two private datasets. Together with the AUC improvements obtained by incorporating textual features in Table~\ref{tab:auc_comparison}, the model attribution analysis and predictive performance results support the effectiveness of the generated features, suggesting that LLM-BlockFE can construct complementary features from long text that are useful for downstream prediction.

\begin{table}[t]
    \centering
    \small
    \setlength{\tabcolsep}{9pt}
    \begin{tabular}{l c c}
        \toprule
        \textbf{Dataset} &
        \textbf{Mean $|\mathrm{SHAP}|$} &
        \textbf{Contribution Ratio} \\
        \midrule
        Remark       & 0.3900 & 41.4\% \\
        Report      & 0.0287 & 61.1\% \\
        CFPB-Timely  & 0.0235 & 9.5\%  \\    
        GossipCop    & 0.6477 & 19.4\% \\
        \bottomrule
    \end{tabular}
    \caption{SHAP-based contribution of the features generated by LLM-BlockFE across datasets.}
    \label{tab:feature_contribution}
\end{table}

\begin{figure}[t]
    \includegraphics[width=\linewidth]{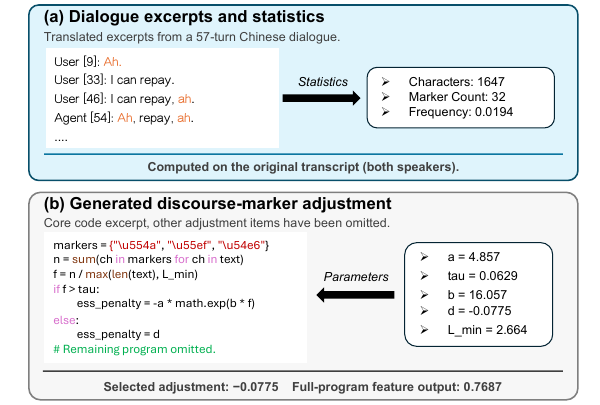}  
    \caption{Case study of a deployed feature program on a real customer-service dialogue.}
    \label{fig:casestudy}
\end{figure}

To illustrate how the generated feature is computed, Figure~\ref{fig:casestudy} presents a dialogue excerpt, statistics for the full text, and the corresponding code snippet. The complete dialogue contains 57 turns and 1,647 characters across both speakers, with 32 occurrences of the specified discourse particles, yielding a normalized frequency of 0.0194. Rather than applying an adjustment directly based on the occurrence count, the code first divides the count by the text length and then compares the resulting frequency with a threshold. Because this sample's frequency is below the threshold of 0.0629, the program follows the default branch and produces an adjustment of $-0.0775$, without activating the exponential adjustment branch. Thus, repeated occurrences of discourse particles do not necessarily trigger a stronger negative adjustment. Their effect depends on their frequency relative to the length of the full text. This example makes the relationship between text statistics, conditional logic, and the local numerical adjustment directly traceable.

\section{Conclusion}

We propose LLM-BlockFE, which formulates feature construction from unstructured long text as an offline search for executable programs. The method builds feature programs by incrementally appending and freezing code blocks. It combines depth-calibrated, credit-guided backtracking with multi-trajectory exploration to reduce dependence on early decisions and limit redundant exploration. The generated programs can be used directly for online feature extraction without further LLM calls, decoupling offline feature discovery from online computation. Experiments show that LLM-BlockFE provides effective complementary features for structured prediction models and achieves the highest AUC in the main experiments. Ablation and generalization experiments further support the contributions of the individual design components and the applicability of the generated features across different downstream models. In five real-world financial risk management applications, post-deployment monitoring shows KS improvements of 0.02 to 1.56 percentage points over existing manually designed strategies. These results suggest that generating inspectable, executable feature programs offline offers a viable approach to leveraging long text information in industrial prediction systems.

\bibliographystyle{ACM-Reference-Format}
\bibliography{sample-base}

@inproceedings{dai2026nsr,
  title={{NSR}-{B}oost: A Neuro-Symbolic Residual Boosting Framework for Industrial Legacy Models},
  author={Dai, Ziming and Ma, Dabiao and Tong, Jinle and Han, Mengyuan and Yang, Jian and Liu, Hongtao and Fei, Haojun and Yang, Qing},
  booktitle={Proceedings of the 32nd ACM SIGKDD Conference on Knowledge Discovery and Data Mining V. 2},
  pages={7129--7140},
  year={2026}
}

@article{abhyankar2025llm,
  title={{LLM-FE}: Automated feature engineering for tabular data with llms as evolutionary optimizers},
  author={Abhyankar, Nikhil and Shojaee, Parshin and Reddy, Chandan K},
  journal={arXiv preprint arXiv:2503.14434},
  year={2025}
}

@article{hollmann2023large,
  title={Large language models for automated data science: Introducing caafe for context-aware automated feature engineering},
  author={Hollmann, Noah and M{\"u}ller, Samuel and Hutter, Frank},
  journal={Advances in Neural Information Processing Systems},
  volume={36},
  pages={44753--44775},
  year={2023}
}

@inproceedings{kim2024discovering,
  title={Discovering Features with Synergistic Interactions in Multiple Views},
  author={Kim, Chohee and Van Der Schaar, Mihaela and Lee, Changhee},
  booktitle={ICML},
  pages={24562--24583},
  year={2024}
}

@inproceedings{malberg2024felix,
  title={{FELIX}: Automatic and interpretable feature engineering using {LLM}s},
  author={Malberg, Simon and Mosca, Edoardo and Groh, Georg},
  booktitle={Joint European Conference on Machine Learning and Knowledge Discovery in Databases},
  pages={230--246},
  year={2024},
  organization={Springer}
}

@article{nam2024optimized,
  title={Optimized feature generation for tabular data via llms with decision tree reasoning},
  author={Nam, Jaehyun and Kim, Kyuyoung and Oh, Seunghyuk and Tack, Jihoon and Kim, Jaehyung and Shin, Jinwoo},
  journal={Advances in neural information processing systems},
  volume={37},
  pages={92352--92380},
  year={2024}
}

@article{burghardt2026famose,
  title={{FAMOSE}: A ReAct Approach to Automated Feature Discovery},
  author={Burghardt, Keith and Liu, Jienan and Sakib, Sadman and Hao, Yuning and Li, Bo},
  journal={arXiv preprint arXiv:2602.17641},
  year={2026}
}

@inproceedings{agarwal2022hierarchical,
  title={Hierarchical Shrinkage: Improving the accuracy and interpretability of tree-based models},
  author={Agarwal, Abhineet and Tan, Yan Shuo and Ronen, Omer and Singh, Chandan and Yu, Bin},
  booktitle={International Conference on Machine Learning},
  pages={111--135},
  year={2022},
  organization={PMLR}
}

@article{yao2023tree,
  title={Tree of thoughts: Deliberate problem solving with large language models},
  author={Yao, Shunyu and Yu, Dian and Zhao, Jeffrey and Shafran, Izhak and Griffiths, Tom and Cao, Yuan and Narasimhan, Karthik},
  journal={Advances in neural information processing systems},
  volume={36},
  pages={11809--11822},
  year={2023}
}

@inproceedings{parsert2024reinforcement,
  title={Reinforcement learning and data-generation for syntax-guided synthesis},
  author={Parsert, Julian and Polgreen, Elizabeth},
  booktitle={Proceedings of the AAAI Conference on Artificial Intelligence},
  volume={38},
  number={9},
  pages={10670--10678},
  year={2024}
}

@article{garthwaite2024locally,
  title={Locally correct confidence intervals for a binomial proportion: A new criteria for an interval estimator},
  author={Garthwaite, Paul H and Moustafa, Maha W and Elfadaly, Fadlalla G},
  journal={Scandinavian Journal of Statistics},
  volume={51},
  number={1},
  pages={220--244},
  year={2024},
  publisher={Wiley Online Library}
}

@article{beltagy2020longformer,
  title={Longformer: The long-document transformer},
  author={Beltagy, Iz and Peters, Matthew E and Cohan, Arman},
  journal={arXiv preprint arXiv:2004.05150},
  year={2020}
}

@article{zaheer2020big,
  title={Big bird: Transformers for longer sequences},
  author={Zaheer, Manzil and Guruganesh, Guru and Dubey, Kumar Avinava and Ainslie, Joshua and Alberti, Chris and Ontanon, Santiago and Pham, Philip and Ravula, Anirudh and Wang, Qifan and Yang, Li and others},
  journal={Advances in neural information processing systems},
  volume={33},
  pages={17283--17297},
  year={2020}
}

@inproceedings{zhang2023openfe,
  title={Open{FE}: Automated feature generation with expert-level performance},
  author={Zhang, Tianping and Zhang, Zheyu Aqa and Fan, Zhiyuan and Luo, Haoyan and Liu, Fengyuan and Liu, Qian and Cao, Wei and Jian, Li},
  booktitle={International Conference on Machine Learning},
  pages={41880--41901},
  year={2023},
  organization={PMLR}
}

@article{balek2025llm,
  title={{LLM}-based feature generation from text for interpretable machine learning},
  author={Balek, Vojt{\v{e}}ch and S{\`y}kora, Luk{\'a}{\v{s}} and Sklen{\'a}k, Vil{\'e}m and Kliegr, Tom{\'a}{\v{s}}},
  journal={Machine Learning},
  volume={114},
  number={11},
  pages={241},
  year={2025},
  publisher={Springer}
}

@article{barlier2026llms,
  title={{LLM}s as Feature Engineers for Text-and-Tabular Prediction},
  author={Barlier, Merwan and Skrlj, Blaz},
  journal={arXiv preprint arXiv:2609.21894},
  year={2026}
}

@inproceedings{chen2016xgboost,
  title={Xgboost: A scalable tree boosting system},
  author={Chen, Tianqi and Guestrin, Carlos},
  booktitle={Proceedings of the 22nd acm sigkdd international conference on knowledge discovery and data mining},
  pages={785--794},
  year={2016}
}

@article{ke2017lightgbm,
  title={Lightgbm: A highly efficient gradient boosting decision tree},
  author={Ke, Guolin and Meng, Qi and Finley, Thomas and Wang, Taifeng and Chen, Wei and Ma, Weidong and Ye, Qiwei and Liu, Tie-Yan},
  journal={Advances in neural information processing systems},
  volume={30},
  year={2017}
}

@article{jiang2023financial,
  title={Financial distress prediction using the Q\&A text of online interactive platforms},
  author={Jiang, Cuiqing and Ma, Lan and Wang, Zhao and Chen, Bo},
  journal={Electronic Commerce Research and Applications},
  volume={61},
  pages={101292},
  year={2023},
  publisher={Elsevier}
}

@article{fujii2022extraction,
  title={Extraction and classification of risk-related sentences from securities reports},
  author={Fujii, Motomasa and Sakaji, Hiroki and Masuyama, Shigeru and Sasaki, Hajime},
  journal={International Journal of Information Management Data Insights},
  volume={2},
  number={2},
  pages={100096},
  year={2022},
  publisher={Elsevier}
}

@article{novikov2025alphaevolve,
  title={Alphaevolve: A coding agent for scientific and algorithmic discovery},
  author={Novikov, Alexander and V{\~u}, Ng{\^a}n and Eisenberger, Marvin and Dupont, Emilien and Huang, Po-Sen and Wagner, Adam Zsolt and Shirobokov, Sergey and Kozlovskii, Borislav and Ruiz, Francisco JR and Mehrabian, Abbas and others},
  journal={arXiv preprint arXiv:2506.13131},
  year={2025}
}

@inproceedings{shojaee2025llm,
  title={{LLM-SR}: Scientific equation discovery via programming with large language models},
  author={Shojaee, Parshin and Meidani, Kazem and Gupta, Shashank and Barati Farimani, Amir and Reddy, Chandan},
  booktitle={International Conference on Learning Representations},
  volume={2025},
  pages={16054--16085},
  year={2025}
}

@inproceedings{fialho2010toward,
  title={Toward comparison-based adaptive operator selection},
  author={Fialho, {\'A}lvaro and Schoenauer, Marc and Sebag, Mich{\`e}le},
  booktitle={Proceedings of the 12th annual conference on Genetic and evolutionary computation},
  pages={767--774},
  year={2010}
}

@misc{seed2025seed-oss,
  author={ByteDance Seed Team},
  title={Seed-OSS Open-Source Models},
  year={2025},
  howpublished={\url{https://github.com/ByteDance-Seed/seed-oss}}
}

@article{shu2020fakenewsnet,
  title={Fakenewsnet: A data repository with news content, social context, and spatiotemporal information for studying fake news on social media},
  author={Shu, Kai and Mahudeswaran, Deepak and Wang, Suhang and Lee, Dongwon and Liu, Huan},
  journal={Big data},
  volume={8},
  number={3},
  pages={171--188},
  year={2020},
  publisher={SAGE Publications Sage CA: Los Angeles, CA}
}

@misc{cfpb_consumer_complaints,
  author       = {{Consumer Financial Protection Bureau}},
  title        = {Consumer Complaint Database},
  year         = {2026},
  howpublished = {\url{https://www.consumerfinance.gov/data-research/consumer-complaints/}},
  note         = {Accessed: 2026-09-30}
}

@article{han2025tabular,
  title={Tabular feature discovery with reasoning type exploration},
  author={Han, Sungwon and Park, Sungkyu and Lee, Seungeon},
  journal={arXiv preprint arXiv:2506.20357},
  year={2025}
}

@article{lundberg2017unified,
  title={A unified approach to interpreting model predictions},
  author={Lundberg, Scott M and Lee, Su-In},
  journal={Advances in neural information processing systems},
  volume={30},
  year={2017}
}

@misc{qwen3.5,
    title  = {{Qwen3.5}: Towards Native Multimodal Agents},
    author = {{Qwen Team}},
    month  = {February},
    year   = {2026},
    url    = {https://qwen.ai/blog?id=qwen3.5}
}

@misc{deepseekai2025deepseekr1incentivizingreasoningcapability,
      title={DeepSeek-R1: Incentivizing Reasoning Capability in LLMs via Reinforcement Learning}, 
      author={DeepSeek-AI},
      year={2025},
      eprint={2501.12948},
      archivePrefix={arXiv},
      primaryClass={cs.CL},
      url={https://arxiv.org/abs/2501.12948}, 
}

@article{qi2024comparison,
  title={A comparison of latent semantic analysis and correspondence analysis of document-term matrices},
  author={Qi, Qianqian and Hessen, David J and Deoskar, Tejaswini and van der Heijden, Peter GM},
  journal={Natural Language Engineering},
  volume={30},
  number={4},
  pages={722--752},
  year={2024},
  publisher={Cambridge University Press}
}

\appendix

\section{Appendix}

\subsection{Dataset Details}\label{apx:A.1}
Table~\ref{tab:dataset_statistics} summarizes all datasets used in our evaluation. The public datasets are primarily derived from open platforms such as Kaggle, while the private industrial datasets are provided by Qfin Holdings. Since high-stakes applications such as financial risk control, which are the focus of this paper, are typically formulated as binary classification tasks, our evaluation focuses uniformly on this setting. To meet the training requirements of the downstream base models and rigorously quantify the performance gains achieved by our method, we specifically select and reconstruct multimodal datasets that contain both initial structured attributes and unstructured long text.

Specifically, we construct the CFPB-Timely dataset from the publicly available CFPB Consumer Complaint Database. This dataset records detailed consumer complaints about various financial products, along with the companies' handling statuses. For our task, we use the ``complaint narrative'' as the unstructured input and whether the company responded in a timely manner as the binary prediction target. We also retain seven basic structured features available when the complaint is received, such as product type, sub-product type, and issue category. For the GossipCop dataset, we use the ``news article text'' as the unstructured long text input and news veracity (real or fake) as the prediction label. In addition, we extract statistical metadata associated with each news article, including the number of images, number of shares, time of dissemination, and intervals between shares, as the dataset's initial structured attributes.
\begin{table}[htbp]
    \centering
    \small
    \begin{tabular}{l c c c c}
        \toprule
        \textbf{Dataset} &
        \makecell[c]{\textbf{Structured}\\\textbf{Features}} &
        \makecell[c]{\textbf{Unstructured}\\\textbf{Long Text}} &
        \textbf{Samples} &
        \textbf{Source} \\
        \midrule
        CFPB-Timely & 7  & 1 & 200,000 & Kaggle \\
        GossipCop   & 7  & 1 & 6,344   & Kaggle \\
        Remark      & 32 & 1 & 100,000 & Qfin   \\
        Report     & 5  & 1 & 100,000 & Qfin   \\
        \bottomrule
    \end{tabular}
    \caption{Dataset statistics.}
    \label{tab:dataset_statistics}
\end{table}

Furthermore, to evaluate the model in real-world credit risk-control applications, we include two private datasets. The Report dataset consists of call records between the company's customer service agents and users. The Remark dataset extracts remarks about target users made by their contacts by parsing authorized address books and using data shared by third-party anti-fraud alliances. Both datasets contain de-identified text and semi-structured records, enabling a comprehensive evaluation of the model's feature extraction capabilities on complex industrial data.

Our dataset selection spans settings from small to large scale. With varied feature dimensionalities and sample sizes, this diverse collection represents a broad range of real-world scenarios and provides a robust framework for evaluating the proposed method.

\subsection{Hyperparameter Settings}\label{apx:a2}

Table~\ref{tab:hyperparameters} reports the hyperparameter settings used by LLM-BlockFE. We use a window length of 12, a prediction horizon of 3 attempts, a prior strength of 4, a rollback threshold of 0.5, and a default success rate of 0.1.

\begin{table}[htbp]
    \centering
    \small
    \begin{tabular}{c l c}
        \toprule
        \textbf{Symbol} & \textbf{Description} & \textbf{Value} \\
        \midrule
        $W$ & Window length & 12 \\
        $H$ & Prediction horizon (attempts) & 3 \\
        $\kappa$ & Prior strength & 4 \\
        $\rho$ & Rollback threshold & 0.5 \\
        $p_0$ & Default success rate & 0.1 \\
        \bottomrule
    \end{tabular}
    \caption{Hyperparameter settings.}
    \label{tab:hyperparameters}
\end{table}

\subsection{Experimental Stability Analysis}

\begin{table*}[t]
    \centering
    \small
    \setlength{\tabcolsep}{4pt}
    \begin{tabular}{l c c c c c c c c c c c}
        \toprule
        \makecell[c]{\textbf{Dataset}} &
        \makecell[c]{\textbf{Base-XGB}} &
        \makecell[c]{\textbf{TextStats}\\\textbf{+XGB}} &
        \makecell[c]{\textbf{TF-IDF}\\\textbf{+LR}} &
        \makecell[c]{\textbf{TF-IDF}\\\textbf{+XGB}} &
        \makecell[c]{\textbf{TF-IDF+SVD}\\\textbf{+XGB}} &
        \makecell[c]{\textbf{Embedding}\\\textbf{+XGB}} &
        \makecell[c]{\textbf{OpenFE}} &
        \makecell[c]{\textbf{CAAFE}} &
        \makecell[c]{\textbf{OCTree}} &
        \makecell[c]{\textbf{LLM-FE}} &
        \makecell[c]{\textbf{Ours}} \\
        \midrule
        Remark
        & $\pm0.0013$ & $\pm0.0026$ & $\pm 0.0000$ & $\pm0.0013$ & $\pm0.0038$ & $\pm0.0031$
        & $\pm0.0028$ & $\pm0.0071$ & $\pm 0.0020$ & $\pm 0.0065$ & $\pm0.0046$ \\
        Report
        & $\pm0.0065$ & $\pm0.0046$ & $\pm 0.0000$ & $\pm0.0043$ & $\pm0.0037$ & $\pm0.0069$
        & $\pm0.0035$ & $\pm 0.0144$ & $\pm 0.0101$ & $\pm 0.0024$ & $\pm0.0088$ \\
        CFPB-Timely
        & $\pm0.0049$ & $\pm0.0023$ & $\pm 0.0000$ & $\pm0.0012$ & $\pm0.0007$ & $\pm0.0016$
        & $\pm0.0050$ & $\pm 0.0063$ & $\pm 0.0034$ & $\pm 0.0051$ & $\pm0.0062$ \\
        GossipCop
        & $\pm0.0008$ & $\pm0.0013$ & $\pm 0.0000$ & $\pm0.0017$ & $\pm0.0009$ & $\pm0.0018$
        & $\pm0.0014$ & $\pm 0.0046$ & $\pm 0.0019$ & $\pm 0.0046$ & $\pm 0.0063$ \\
        \bottomrule
    \end{tabular}
    \caption{The standard deviation of AUC values obtained from five runs on different datasets. }
    \label{tab:auc_std}
\end{table*}

We run each method independently five times on each dataset and compute the standard deviation of test AUC to assess experimental stability. As shown in Table~\ref{tab:auc_std}, the standard deviations of LLM-BlockFE on Remark, Report, CFPB-Timely, and GossipCop are 0.0046, 0.0088, 0.0062, and 0.0063, respectively. The results show that variability across runs differs across datasets, with relatively greater variability on Report. Overall, its standard deviations are broadly comparable to those of the LLM-based baselines, indicating that the observed performance is not attributable to a single run.

\end{document}